\documentclass[a4paper,12pt]{article}

\usepackage{url}
\usepackage[dvipsnames]{xcolor}
\usepackage[utf8]{inputenc} 
\usepackage[T1]{fontenc}
\usepackage[a4paper, total={7in, 9in}]{geometry}
\usepackage[dvipsnames]{xcolor}
\usepackage{graphicx}
\usepackage{amscd,amsmath,amssymb,amsfonts,latexsym,mathrsfs,amsthm,mathtools}	
\usepackage{bm}                         % sloping bold
\usepackage{isomath}                    % fancy italicised vectors
\usepackage{subcaption}
\usepackage{cancel} 
\usepackage[toc,page]{appendix} 
\usepackage{titling}

\newcommand{\Exp}{\mathbb{E}}

\definecolor{myblue}{RGB}{0,163,243}

\newcommand{\x}{{\bf x}}
\newtheorem*{theo*}{Theorem}

\newcommand{\norm}[1]{\left\lVert#1\right\rVert}

\begin{document}

\setlength{\droptitle}{-7em} 
\title{Optimality in Noisy Importance Sampling} % \\
%With Applications to Reinforcement Learning}}
%\subtitle{With Applications to Reinforcement Learning}
\date{\vspace{-5ex}}

\author{Fernando Llorente$^*$, Luca Martino$^\star$, Jesse Read$^\dagger$, David Delgado--G\'omez$^*$ \\
	{\small $^*$ Universidad Carlos III de Madrid, Legan{\'e}s, Spain. } \\
	{\small $^\star$ Universidad Rey Juan Carlos, Fuenlabrada, Spain. } \\
	{\small $^\dagger$ École Polytechnique, Palaiseau, France. }
}

\maketitle

%\vspace{-1cm}

\begin{abstract}
	Many applications in signal processing and machine learning require the study of probability density functions (pdfs) that can only be accessed through noisy evaluations.	
%	There exist Monte Carlo methods that can cope with these noisy evaluations without compromising its convergence.
In this work,  we analyze the noisy importance sampling (IS), i.e., IS working with noisy evaluations of the target density. 
We present the general framework and derive optimal proposal densities for noisy IS estimators. 
	The optimal proposals incorporate the information of the variance of the noisy realizations, proposing points in regions where the noise power is higher.
	We also compare the use of the optimal proposals with previous optimality approaches considered in a noisy IS framework.
\end{abstract}

\section{Introduction}
\label{sec-intro}

%%%%%%%%%%%%%%%%%%%
%\subsection{Application scenarios}\label{sec_appEsc} 
%%%%%%%%%%%%%%%%%%%%

%{\bf REDUCIR EN UNOS PARRAFOS (1 o 2 parrafos) - decir tambien algo de surrogate models ....tambien for approximate integrals...}

A wide range of modern applications, especially in Bayesian inference framework \cite{luengo2020survey}, require the study of probability density functions (pdfs) which can be evaluated stochastically, i.e., only noisy evaluations can be obtained  \cite{fearnhead2010random,tran2013importance,acerbi2020variational,Llorente2021arxiv}. For instance, this is the case of the  pseudo-marginal approaches and doubly intractable posteriors \cite{andrieu2009pseudo,park2018bayesian},  approximate Bayesian computation (ABC) and likelihood-free schemes  \cite{price2018bayesian,gutmann2016bayesian}, where the target density cannot be computed in closed-form. 
\newline
The noisy scenario also appears naturally when mini-batches of data are employed instead of considering the complete likelihood of huge amounts of data \cite{bardenet2017markov,quiroz2018speeding}. More recently, the analysis of noisy functions of densities is required in  reinforcement learning (RL), especially in  direct policy search which is an important branch of RL, with applications  in robotics \cite{deisenroth2013survey,handful}.
The topic of inference in noisy settings (or where a function is known with a certain degree of uncertainty) is also of interest in the inverse problem literature, such as in the calibration of expensive computer codes \cite{duncan2021ensemble,bliznyuk2008bayesian}. This is also the case when the construction of an {\it emulator} is considered,  as a surrogate model  \cite{acerbi2020variational,svendsen2020active,llorente2020adaptive}. 
% where emulators of the forward model are built fromvaluations corrupted with noise, and then these emulators are subsequently sampled using MCMC algorithms.
%Noisy likelihood evaluations are also considered for building surrogates, and then use them in order to obtain a variational approximations to the posterior \cite{acerbi2020variational}.

In this work, we study the importance sampling (IS) scheme under noisy evaluations of the target pdf. The noisy IS scenario has been already analyzed in the literature \cite{fearnhead2010random,tran2013importance,fearnhead2008particle}. In the context of optimization, some theoretical results can be found \cite{akyildiz2017adaptive}. In the sequential framework, IS schemes with random weights can be found and have been studied in different works \cite{fearnhead2010random,crisan2018nested,chopin2013smc2,Martino15PF}.  
We provide the optimal proposal densities for different noisy scenarios,  including also the case of integrals involving vector-valued functions.   Moreover, we discuss the convergence and variance of the estimators in a  general setting.  We consider a different approach with respect to other studies in the literature \cite{tran2013importance,doucet2015efficient}. In those works, the authors analyzed the trade-off between decreasing the noise power (by increasing the number of auxiliary samples) and increasing the total number of samples in the IS estimators. Here, this information is encompassed within the optimal proposal density, which plays a similar role to an acquisition function in active learning \cite{llorente2020adaptive,svendsen2020active}. % remarking the regions which need more sampling.
This is information is relevant, especially if the noisy evaluations are also costly to obtain.

%{\color{red}
%%The presence of noise in the target evaluation must be addressed since:
%%
%%-- (i) when noise is unbiased, the resulting noisy IS estimators are less efficient, i.e., they have greater variance $\rightarrow$ optimal proposal depends on $s(\x)^2$, the spatial distribution of noise strength
%%
%%-- (ii) when noise is arbitrary, the resulting noisy IS estimators converge to expectations under $m(\x) = \mathbb{E}[\widetilde{\pi}(\x)]$, 
%
%In this paper, we provide
%
%
%
%
%%Para darle diferencia respecto a esos papers, podemos discutir lo de la proposal optimal. En Tran et al se centran en discutir el trade-off entre noiseness (number of auxiliary samples) y number of particles
%}

%The Markov Chain Monte Carlo (MCMC) are well-known Monte Carlo (MC) methodologies, which have become very popular in signal processing, statistics and machine learning during the past decades, in order to perform Bayesian inference and stochastic optimization \cite{Candy09,Fitzgerald01,Robert04}. They generate a Markov chain which converges to the desired stationary probability density function (pdf). 
%
%\dots
%
%The remaining of the paper is organized as follows. The problem statement is described in Section \ref{PSsect}. The proposed technique is introduced in Section \ref{GenOUTLINE}. Section \ref{Simu1} contains the numerical experiments. We conclude with a brief summary in Section \ref{sec:conclusions}.

%%%%%%%%%%%%%%%%%%%
\section{Background}
%%%%%%%%%%%%%%%%%%%

%%
\subsection{Bayesian inference}
\label{PSsect}

In many applications, we aim at inferring a variable of interest given a set of observations or measurements.
Let us denote the variable of interest by ${\bf x}\in \mathcal{D}\subseteq \mathbb{R}^{d_x}$, and let ${\bf y}\in \mathbb{R}^{d_y}$ be the observed data.
The posterior pdf is then
\begin{equation}
\bar{p}({\bf x}| {\bf y})= \frac{\ell({\bf y}|{\bf x}) g({\bf x})}{Z({\bf y})},
\label{eq:posterior}
\end{equation}
where $\ell({\bf y}|{\bf x})$ is the likelihood function, $g({\bf x})$ is the prior pdf, and $Z({\bf y})$ is the model evidence (a.k.a. marginal likelihood) which is a useful quantity in model selection problems \cite{ourRev}. For simplicity, in the following, we skip the dependence on ${\bf y}$ in $\bar{p}({\bf x})=\bar{p}({\bf x}| {\bf y})$ and $Z=Z({\bf y})$.
Generally, $Z$ is unknown, so we are able to evaluate the unnormalized target function, i.e., the numerator on the right hand side of Eq. \eqref{eq:posterior},
\begin{equation}
p({\bf x})=\ell({\bf y}|{\bf x}) g({\bf x}).
\label{eq:target}
\end{equation}
The analytical study of the posterior density $\bar{p}({\bf x}) \propto p({\bf x})$ is unfeasible, so that numerical approximations are required \cite{Robert04,luengo2020survey}.

\subsection{Noisy framework}

Generally, we desire to approximate  the unnormalized density $p(\x),\ \x\in \mathcal{X}\subset \mathbb{R}^d$, and  the corresponding normalizing constant $Z$, using Monte Carlo methods. 
The  unnormalized density $p(\x)$ can represent a posterior density in a Bayesian inference problem, as described above.
We assume that, for any $\x$, we cannot evaluate $p(\x)$ exactly, but  we only have access to a related noisy realization. Moreover, in many applications, obtaining such a noisy realization may be expensive. 
Hence, analyzing in which $\x$ we require a noisy realization of $p(\x)$ is an important problem, which is related to the concept of {\it optimality} that we consider below.   
%Typically, this occurs in applications where the function of interest $p(\x)$ is intractable or expensive to evaluate.
\newline 
In the following, we introduce a concise  mathematical formalization of the noisy scenario. This simple framework contains real application scenarios, such as latent variable models \cite{tran2013importance} (see example 4 in Sect. \ref{sec_qopt_Z}), likelihood-free inference setting \cite{price2018bayesian}, doubly intractable posteriors \cite{park2018bayesian}, mini batch-based inference \cite{bardenet2017markov}.
%In the following, we formalize the idea of related noisy realization, as a generic function of the true evaluation and some stochastic error.}
More specifically,  we assume to have access to a noisy realization related to $p(\x)$, i.e.,
%{\color{red}
\begin{equation}
\widetilde{m}(\x) = H(p(\x),\epsilon),
\end{equation}
%}
where $H$ is a non-linear transformation involving $p(\x)$ and $\epsilon$, that is some noise perturbation.
Thus, {\it for a fixed value $\x$}, $\widetilde{m}(\x)$ is a random variable with
\begin{equation}
 m(\x)=\mathbb{E}[\widetilde{m}(\x)], \quad  s(\x)^2=\text{Var}[\widetilde{m}(\x)],
\end{equation}
for some {\em mean function}, $m(\x)$, and {\em variance function}, $s(\x)^2$. 
The assumption that $\widetilde{m}(\x)$  must be strictly positive is important in practice \cite{fearnhead2010random,doucet2015efficient}. 
%{\color{red}decir que $\widetilde{m}(\x)$ tiene ser strictly positive }
\newline
\newline
{\bf Noise power.} In some applications, it is also possible to control the noise power $s(\x)^2$, for instance by adding/removing data to the mini-batches (e.g., in the context of Big Data) \cite{bardenet2017markov}, increasing the number of auxiliary samples in latent variables models \cite{andrieu2009pseudo},  or interacting with an environment over longer/shorter periods of time (e.g., in reinforcement learning) \cite{deisenroth2013survey}.
\newline
\newline
{\bf Unbiased scenario and related cases.} The scenario where $m(\x)=p(\x)$ appears naturally in some applications (such as in the estimation of latent variable and stochastic volatility models in statistics \cite{tran2013importance,andrieu2010particle}; or in the context of optimal filtering of partially observed stochastic processes \cite{fearnhead2008particle}), or it is often assumed as a pre-established condition by the authors  \cite{tran2013importance,fearnhead2010random}.  
%This is the case of the so-called pseudo marginal approaches \cite{Pseudo-Marginal}. 
In some other scenarios, the noisy realizations are known to be unbiased estimates of some transformation of $p(\x)$, e.g., of $\log p(\x)$ \cite{jarvenpaa2021parallel,drovandi2018accelerating}.  %In this situation, it is possible to work in a transformed domain
This situation can be encompassed by the following special case. If we consider an additive perturbation, 
\begin{align}\label{eq_con_ruido_adi}
%	{\color{red}
\widetilde{m}(\x) = G\left(p(\x)\right)+ \epsilon, \qquad \mbox{with $\Exp[\epsilon]=0$,}
%	}
\end{align}
we have $m(\x)=\Exp[\widetilde{m}(\x)]=G\left(p(\x)\right)$, where $G(\cdot):\mathbb{R}\rightarrow \mathbb{R}$. If $G$ is known and invertible,  we have $p(\x)=G^{-1}\left(m(\x)\right)$. %\cite{jarvenpaa2021parallel,drovandi2018accelerating}.  
%	Generally, transforming $\widetilde{m}(\x)$ into an unbiased realization of $p(\x)$ is not straightforward, since $\Exp[G^{-1}(\widetilde{m}(\x))]\neq p(\x)$. However, there are cases such as $\widetilde{m}(\x) = \log p(\x) + \epsilon$, where we can take
%	$\widetilde{p}(\x) = e^{\widetilde{m}(\x)}$ which fulfills $\Exp[\widetilde{p}(\x)] \propto p(\x)$ .
\newline
Generally, we can state that $m(\x)$ always contains statistical information related to $p(\x)$. The subsequent use of $m(\x)$ depends on the specific application. Thus, we study the mean function $m(\x)$.
%\newline
%{\color{red} add frase para enganchar... y (a lo mejor arriba) hablar de $m({\bf x})$ y $p(\bf x)$ (quizas remark) ...}
Hence, our goal is to approximate efficiently integrals involving $m(\x)$, i.e.,
\begin{equation}\label{Goal_in_Integral}
{\bf I}=\frac{1}{\bar{Z}}\int_{\mathcal{X}} {\bf f}(\x) m(\x) d\x, \quad \bar{Z}=\int_{\mathcal{X}}  m(\x) d\x, 
\end{equation}
where ${\bf f}(\x):  \mathcal{X}\rightarrow \mathbb{R}^{d_f}$ and ${\bf I}=[I_1,\dots,I_{d_f}]^\top\in\mathbb{R}^{d_f}$ denotes the vector of integrals of interest.
%	with $I_i=\frac{1}{\bar{Z}}\int_\mathcal{X}f_i(\x)d\x$
Note that, in the unbiased case $m(\x)=p(\x)$, we have $ \bar{Z}=Z$.  An integral involving $m({\bf x})$ can be approximated employing a cloud of random samples using the noisy realizations $\widetilde{m}(\x)$ via Monte Carlo methods.

\section{Noisy Importance Sampling}
%%%%%%%%%%%%%%%%%%%%%%%%%%%%%%%%%%%%%

In a non-noisy IS scheme, a set of samples is drawn from a proposal density $q(\x)$. Then each sample is weighted according to the ratio $\frac{p(\x)}{q(\x)}$.  A noisy version of importance sampling can be obtained when we substitute the evaluations of $p(\x)$ with noisy realizations of $\widetilde{m}(\x)$.
See Table \ref{tab_noisyIS} and note that the importance weights $w_n$ in Eq. \eqref{eq_ISweightComp} are computed using the noisy realizations. Below, we show that 
\vspace{-0.5cm}
\begin{align}
\widehat{Z}=\frac{1}{N}\sum_{n=1}^N w_n, 
\end{align}
 is an unbiased estimator of $\bar{Z}$, and
\begin{align}
%\label{eq_std_IS}
\widehat{{\bf I}}_\text{std}=\frac{1}{N \bar{Z}} \sum_{n=1}^N w_n {\bf f}(\x_n),  \qquad 
\widehat{{\bf I}}_\text{self}=\frac{1}{N \widehat{Z}} \sum_{n=1}^N w_n {\bf f}(\x_n), \label{AquiSelfNormISnoisy}
\end{align}
 are consistent estimators of ${\bf I}$. The estimator $\widehat{{\bf I}}_\text{std}$ requires the knowledge of $\bar{Z}$, that is not needed in the so-called self-normalized estimator, $\widehat{{\bf I}}_\text{self}$.

%%%%%%%%%%%%%
\begin{table}[!h]
	\centering
	\caption{Noisy importance sampling algorithm}
	{\footnotesize
		\begin{tabular}{|p{0.95\columnwidth}|}
			\hline
			\begin{itemize}
				\item[1.] \textbf{Inputs:} Proposal distribution  $q(\x)$.
				
				\item[2.]\label{} For $n=1,\dots,N$:
				\begin{enumerate}
					\item[(a)]  Sample $\x_n \sim q(\x)$ and obtain one realization $\widetilde{m}(\x_n)$.

					\item[(b)] Compute 
					\begin{align}\label{eq_ISweightComp}
					w_n = \frac{\widetilde{m}(\x_n)}{q(\x_n)}
					\end{align}
				\end{enumerate}	
				
				%\item[3] Compute normalized weights: $\bar{w}_n = \frac{w_n}{\sum_{j=1}^Nw_j},\ j=1,\dots,N$.

				\item[4] \textbf{Outputs:} the weighted samples $\{\x_{n},w_n\}_{n=1}^N$.
			\end{itemize}\\
			\hline
		\end{tabular}	
	}
	\label{tab_noisyIS}
	\vspace{0.2cm}
\end{table}

%%%
%\subsection{Convergence}
%%

\begin{theo*}\label{teo1_noisyMH}
	%Under certain conditions,
	 The estimators above constructed from the output of noisy IS converge to expectations under $m(\x)$. 
	 More specifically, we have $\widehat{Z}$ and $\widehat{{\bf I}}_\text{std}$ are unbiased estimators of $\bar{Z}$ and ${\bf I}$ respectively, and  $\widehat{{\bf I}}_\text{self}$ is a consistent estimator of ${\bf I}$.
	 Moreover, these estimators have higher variance than their non-noisy counterparts.

%	We show that an IS estimator built with noisy realizations $\widetilde{m}(\x)$, converges to expectations w.r.t.\ $m(\x)$.
%	Let $q(\x)$ denote a proposal pdf, and let 
%	\begin{align}
%	\widetilde{Z} = \frac{1}{N}\sum_{i=1}^{N}\frac{\widetilde{m}(\x_i)}{q(\x_i)}=\frac{1}{N}\sum_{i=1}^{N}\widetilde{w}_i,
%	\end{align}
%	be the IS estimator built with noisy realizations, where $\widetilde{w}_i=\frac{\widetilde{m}(\x_i)}{q(\x_i)}$ are the noisy weights, and $\{\x_i\}_{i=1}^N$ are iid samples from $q$.
%	The non-noisy IS estimator
%	\begin{align}
%	\widehat{Z} = \frac{1}{N}\sum_{i=1}^{N}\frac{m(\x_i)}{q(\x_i)}=\frac{1}{N}\sum_{i=1}^{N}{w}_i,
%	\end{align} 
%	is an unbiased estimator of $\bar{Z}= \int m(\x)d\x$, i.e., $\Exp[\widehat{Z}]=Z$, converging as $N \to \infty$ at rate $\frac{1}{N}$.
%	We aim to show that $\widetilde{Z}$ is also an unbiased estimator of $\bar{Z}$, with greater variance than $\widehat{Z}$, but the same convergence speed, i.e., its variance decreases at $\frac{1}{N}$ rate. 
%	%In order to show that $\widetilde{I}$ converges to $I$ as $N\to\infty$, it is sufficient to show that (i) $\mathbb{E}[\widetilde{I}]=I$, and (ii) $\text{Var}[\widetilde{I}]\propto \frac{1}{N}$. 
	
	\begin{proof}
	Here, we provide a simple proof of convergence by  applying iterated conditional expectations.
	Equivalently, the correctness of the approach can be proved by using an extended space view (see, e.g., \cite{fearnhead2008particle,tran2013importance}).
	\newline
	Let ${\bf x}_{1:N}=[\x_1,\dots,\x_N]$ denote the $N$ samples from $q$.
	By the law of total expectation, we have that $\Exp[\widehat{Z}] = \Exp\left[\Exp[\widehat{Z}|{\bf x}_{1:N}]\right]$. In the inner expectation, we use the fact the ${w}_i$'s are i.i.d., hence
	\begin{align*}
	\Exp[\widehat{Z}|{\bf x}_{1:N}] = \frac{1}{N}\sum_{i=1}^{N}\Exp[w_i|\x_i]&= \frac{1}{N}\sum_{i=1}^{N}\frac{1}{q(\x_i)}\Exp[\widetilde{m}(\x_i)|\x_i]=  \frac{1}{N}\sum_{i=1}^{N}\frac{m(\x_i)}{q(\x_i)}=\widetilde{Z},
	\end{align*}
	where $\widetilde{Z}$ is the non-noisy IS estimator of $\bar{Z}$, which is also unbiased, i.e.,
	\begin{align*}
	\Exp[\widehat{Z}] = \Exp\left[\Exp[\widehat{Z}|{\bf x}_{1:N}]\right] = \Exp[\widetilde{Z}] = \bar{Z}.
	\end{align*}
	Therefore, $\widehat{Z}$ is an unbiased estimator of $\bar{Z}= \int_{\mathcal{X}} m(\x)d\x$, i.e., $\Exp[\widehat{Z}]=\bar{Z}$.  Moreover, we show below that  $\text{Var}[\widehat{Z}] $ decreases to zero as $N \to \infty$. Hence, $\widehat{Z}$  is a consistent estimator of $\bar{Z}$.  %at rate $\frac{1}{N}$.
	Now, with the same arguments, we can  prove that the estimator $\widehat{{\bf E}} = \frac{1}{N}\sum_{i=1}^{N}\frac{\widetilde{m}(\x){\bf f}(\x)}{q(\x)}$ is also unbiased and converges to ${\bf E} = \int_{\mathcal{X}} {\bf f}(\x)m(\x)d\x$. 
	Thus, both the estimator $\widehat{{\bf I}}_\text{std}$, and the ratio 
	$$
	\widehat{{\bf I}}=\frac{1}{\widehat{Z}} \widehat{{\bf E}} = \frac{1}{\sum_{j=1}^N w_j}\sum_{i=1}^N w_i {\bf f}(\x_i),
	$$
	which is the noisy self-normalized IS estimator $\widehat{{\bf I}}_\text{self}$ in Eq. \eqref{AquiSelfNormISnoisy}, are consistent estimators of 
	$$
	{\bf I}=\frac{\int_{ \mathcal{X}}  {\bf f}(\x)m(\x)d\x}{\int_{\mathcal{X}} m(\x)d\x}=\frac{1}{\bar{Z}}\int_{\mathcal{X}} {\bf f}(\x)m(\x)d\x,
	$$ 
	given in Eq. \eqref{Goal_in_Integral}.
%	\newline

\end{proof}
\end{theo*}

%\newline
\noindent{\bf Variance of $\widehat{Z}$.} By the law of total variance, we have that
	\begin{align*}
	\text{Var}[\widehat{Z}] = \Exp\left[\text{Var}[\widehat{Z}|{\bf x}_{1:N}]\right] + \text{Var}\left[\Exp[\widehat{Z}|{\bf x}_{1:N}]\right].
	\end{align*}
In a non-noisy scenario, i.e., in a non-noisy IS setting, the first term is null.
Using the fact that $\widehat{Z}$ is unbiased, we have that the second term is
	\begin{align*}
	\text{Var}\left[\Exp[\widehat{Z}|{\bf x}_{1:N}]\right] = \text{Var}[\widetilde{Z}] = \mathcal{O}\left(1/N\right).
	\end{align*}
Regarding the first term, we have
	\begin{align*}
	\text{Var}[\widehat{Z}|{\bf x}_{1:N}] =\frac{1}{N^2}\sum_{i=1}^{N}\text{Var}[{w}_i|\x_i]&=\frac{1}{N^2}\sum_{i=1}^{N}\frac{1}{q(\x_i)^2}\text{Var}[\widetilde{m}(\x_i)|\x_i]=\frac{1}{N^2}\sum_{i=1}^{N}\frac{s(\x_i)^2}{q(\x_i)^2}.
	\end{align*}
Assuming that $\frac{s(\x)^2}{q(\x)^2}<\infty$ for all $\x$, we have that
	\begin{align*}
	\Exp\left[\text{Var}[\widehat{Z}|{\bf x}_{1:N}]\right] 
	&=\frac{1}{N^2}\sum_{i=1}^{N}\Exp\left[\frac{s(\x_i)^2}{q(\x_i)^2}\right]=\frac{1}{N}\Exp\left[\frac{s(\x)^2}{q(\x)^2}\right], \quad \mbox{ where } \quad \x\sim q(\x).
	\end{align*}
Hence, we finally have that
	\begin{align}\label{eq_var_of_Z}
	\text{Var}[\widehat{Z}] &= \frac{1}{N}\Exp\left[\frac{s(\x)^2}{q(\x)^2}\right] + \text{Var}[\widetilde{Z}] \geq \text{Var}[\widetilde{Z}]. 
	\end{align}
	%{\color{red}From this expression, we can deduce that the variance of $\widehat{Z}$ depends on the mismatch between $q(\x)$ and $s(\x)$.} 
Therefore, $\widehat{Z}$ has a greater variance than $\widetilde{Z}$, but the same convergence speed, i.e., its variance decreases at $\frac{1}{N}$ rate. Proving that $\widehat{{\bf E}}$ has greater variance than its non-noisy version is straightforward.

%%%%%%%%%%%%%%%%%%%%%%%%%%%%%
\section{Optimal Proposal Density in Noisy IS}
%%%%%%%%%%%%%%%%%%%%%%%%%%%%%%

In this section, we derive the optimal proposals for the noisy IS estimators $\widehat{Z}$, $\widehat{{\bf I}}_\text{std}$ and $\widehat{{\bf I}}_\text{self}$.

\subsection{Optimal proposal for $\widehat{Z}$}\label{sec_qopt_Z}

We can rewrite the variance of $\widehat{Z}$ in Eq. \eqref{eq_var_of_Z} as
\begin{align*}
\mbox{Var}[\widehat{Z}] 
%&= \mbox{Var}[\widetilde{Z}] + \frac{1}{N}\mathbb{E}\left[\frac{s(\x)^2}{q(\x)^2}\right] \\
&= \frac{1}{N}\mathbb{E}\left[\frac{m(\x)^2+s(\x)^2}{q(\x)^2}\right] - \frac{1}{N}\bar{Z}^2.
\end{align*}
By Jensen's inequality, the first term is bounded below by
\begin{align*}
	\mathbb{E}\left[\frac{m(\x)^2+s(\x)^2}{q(\x)^2}\right] \geq \left(\mathbb{E}\left[\frac{\sqrt{m(\x)^2+s(\x)^2}}{q(\x)}\right]\right)^2.
\end{align*}
The minimum variance $\mbox{V}_\text{min}=\min_q\mbox{Var}[\widehat{Z}]$ is thus attained at 
 \begin{equation}\label{aquiVaOpt}
 q_\text{opt}(\x)\propto \sqrt{m(\x)^2+s(\x)^2},
\end{equation}
  Note that, for finite $N$, $\mbox{V}_\text{min}$ is always greater than 0, specifically,
\begin{align}\label{eq_Vopt}
	\mbox{V}_\text{min} = \frac{1}{N}\left[\int_{\mathcal{X}} \sqrt{m(\x)^2+s(\x)^2}d\x\right]^2 - \frac{1}{N}\bar{Z}^2.
\end{align}
Hence, differently from the non-noisy setting, in noisy IS the optimal proposal does not provide an estimator with null variance. If $s(\x)=0$ for all $\x$,  then we come back to the non-noisy scenario and  $\mbox{V}_\text{min}=\frac{1}{N}\left[\int_{\mathcal{X}} m(\x)d\x\right]^2 - \frac{1}{N}\bar{Z}^2=0$. Note that the variance of using $q(\x)=\frac{1}{\bar{Z}}m(\x)$ is
\begin{align}\label{eq_Vsub}
	\mbox{V}_\text{sub-opt} 
	&= \frac{\bar{Z}}{N}\int_{\mathcal{X}} \frac{m(\x)^2+s(\x)^2}{m(\x)} d\x - \frac{1}{N}\bar{Z}^2 = \frac{\bar{Z}}{N}\int_{\mathcal{X}} \frac{s(\x)^2}{m(\x)}d\x.
\end{align}
In the following, we show several examples of noise models and their corresponding optimal proposal densities.
\newline
\noindent
{\bf Example 1.} Let us consider a Bernoulli-type noise where $\widetilde{m}(\x) = p_\text{max} \epsilon$, where $\epsilon \sim \mbox{Bernoulli}\left(\frac{p(\x)}{p_\text{max}}\right)$, and $p_\text{max} = \max p(\x)$. Then, we have 
\begin{align*}
m(\x)= p(\x),\qquad s(\x)^2 = p(\x)[p_\text{max}-p(\x)].
\end{align*}
Replacing in Eq. \eqref{aquiVaOpt}, the optimal proposal density in this case is
\begin{align}
q_\text{opt}(\x) \propto p(\x)\sqrt{1 + [p_\text{max}-p(\x)]^2}.
\end{align}
\noindent
{\bf Example 2.}  Let us consider $\widetilde{m}(\x) =|p(\x)+\epsilon|$,  with $\epsilon \sim \mathcal{N}(0,\sigma^2)$. In this scenario, the random variable $\widetilde{m}(\x)$ corresponds to a folded Gaussian random variable.
We have
\begin{align*}
	m(\x) &= \sigma\sqrt{\frac{2}{\pi}}\exp\left(-p^2(\x)/2\sigma^2\right) + p(\x)[1-2\bm{\Phi}(-p(\x)/\sigma)], \\
	s(\x)^2 &= p(\x)^2 + \sigma^2 - m(\x)^2,
\end{align*}
where $\bm{\Phi}(\x)$ is the cumulative function of the standard Gaussian distribution. Then, 
\begin{align}
	q_\text{opt}(\x) \propto \sqrt{p(\x)^2 + \sigma^2}.
\end{align}
%\newline
%\newline
\noindent{\bf Example 3.}  Let us consider a multiplicative noise $\widetilde{m}(\x) = e^\epsilon p(\x)$ with $\mathbb{E}[\epsilon]=0$, hence 
\begin{align*}
m(\x)=p(\x)  \mathbb{E}[e^\epsilon] \propto p(\x), \qquad
s(\x)^2 =p(\x)^2\mbox{Var}[e^\epsilon] .
\end{align*}
 If we denote $A=\mathbb{E}[e^\epsilon] $ and $\sigma^2 = \mbox{Var}[e^\epsilon]$, then  $m(\x)=Ap(\x)$ and $s(\x)^2 =\sigma^2 p^2(\x)$. In this case, the optimal proposal coincides with the optimal one in the non-noisy setting, since 
\begin{align}
q_\text{opt}(\x) \propto \sqrt{A^2p^2(\x)+\sigma^2p^2(\x)} = p(\x)\sqrt{A^2+\sigma^2} \propto p(\x).
\end{align}
\noindent{\bf Example 4.}
In latent variable models, the noisy realization corresponds to the product of $d_y$ independent IS estimators, each built from $R$ auxiliary samples.
%	$$
%	\widehat{p}_N({\bf y}|\x) = \prod_{i=1}^n\widehat{p}_N(y_i|\x),
%	$$
%where 
%	$$
%	\widehat{p}_N(y_i|\x) = \frac{1}{N}\sum_{j=1}^N\frac{p(y_i|\alpha_i^{(j)},\x)p(\alpha_i^{(j)}|\x)}{g(\alpha_i^{(j)})},\quad \alpha_i^{(j)}\sim g(\alpha_i)
%	$$
%with $p(y_i,\alpha_i|\x)=p(y_i|\alpha_i,\x)p(\alpha_i|\x)$ being the conditional likelihood and $g(\alpha_i)$ being some proposal density for $\alpha_i$, the $i$-th latent variable.
With $d_y$ large enough, the distribution of this realization is approximately lognormal, i.e., $$\widetilde{m}(\x) \sim \log\mathcal{N}(\mu(\x),\sigma^2(\x)),$$
%$$
%\log \widehat{p}_N({y}|\x) \sim \log \mathcal{N}(\cdot,\cdot)
%$$
%They make the following assumption $\widehat{\pi}_N(\x) \sim $ where
where $\mu(\x) = \log p(\x) - \frac{\gamma^2(\x)}{2R}$ and $\sigma^2(\x) = \frac{\gamma^2(\x)}{R}$, for some function $\gamma^2(\x)$ \cite{tran2013importance,doucet2015efficient}. 
Equivalently, they write $\widetilde{m}(\x) = p(\x)e^\epsilon$, where $\epsilon \sim  \mathcal{N}(\mu(\x),\sigma^2(\x))$.
Hence,
\begin{align*}
	m(\x) = p(\x), \qquad
	s(\x)^2 = (e^{\gamma^2(\x)/R}-1)p(\x)^2,
\end{align*}
and the optimal proposal is 
\begin{align}
	q_\text{opt}(\x)\propto p(\x)e^{\frac{\gamma^2(\x)}{2R}}.
\end{align}
This example is related with the cases studied in \cite{tran2013importance,doucet2015efficient}.

\subsection{Optimal proposal for $\widehat{{\bf I}}_{std}$ }

We have already seen that the optimal proposal that minimizes the variance of $\widehat{Z}$ is $q_\text{opt}(\x) \propto \sqrt{m(\x)^2 + s(\x)^2}$. 
Let us consider now the estimator $\widehat{{\bf I}}_\text{std}$. Note that this estimator assumes we can evaluate $\bar{Z}=\int_{\mathcal{X}} m(\x)d\x$. 
Since we are considering a vector-valued function, the estimator has $d_f$ components  $\widehat{{\bf I}}_\text{std}=[\widehat{I}_{\text{std},1} \dots \widehat{I}_{\text{std},d_f}]^\top$, and $\text{Var}[\widehat{{\bf I}}_\text{std}]$ corresponds to a $d_f \times d_f$ covariance matrix.
We aim to find the proposal that minimizes the sum of diagonal variances. From the results of the previous section, it is straightforward to show that the variance of the $p$-th component is
\begin{align*}
	\text{Var}[\widehat{I}_{\text{std},p}] 
	&= \text{Var}[\widetilde{I}_{\text{std},p}] + \frac{1}{N\bar{Z}^2}\mathbb{E}\left[\frac{f_p(\x)^2s(\x)^2}{q(\x)^2}\right] \\
	&= \frac{1}{N\bar{Z}^2}\mathbb{E}\left[\frac{f_p(\x)^2(m(\x)^2+s(\x)^2)}{q(\x)^2}\right] - \frac{1}{N\bar{Z}^2}I_p^2,
\end{align*}
where $f_p(\x)$ and $I_p$ are respectively the $p$-th components of ${\bf f}(\x)$ and ${\bf I}$, and $\widetilde{I}_{\text{std},p}$ denotes the non-noisy estimator (i.e. using $m(\x)$ instead of $\widetilde{m}(\x)$). 
Thus,
{\footnotesize$$
\sum_{p=1}^{d_f} \text{Var}[\widehat{I}_{\text{std},p}] = \frac{1}{N\bar{Z}^2}\mathbb{E}\left[\frac{\sum_{p=1}^{d_f} f_p(\x)^2(m(\x)^2+s(\x)^2)}{q(\x)^2}\right] - \frac{1}{N\bar{Z}^2}\sum_{p=1}^{d_f} I_p^2.
$$}
By Jensen's inequality, we have
{\footnotesize
\begin{align*}
	\mathbb{E}\left[\frac{\sum_{p=1}^{d_f} f_p(\x)^2(m(\x)^2+s(\x)^2)}{q(\x)^2}\right] 
%	&\geq \left(\mathbb{E}\left[\frac{\sqrt{\sum_{p=1}^\chi f_p(\x)^2(m(\x)^2+s(\x)^2)}}{q(\x)}\right]\right)^2 \\
&\geq\left(\mathbb{E}\left[\frac{\sqrt{m(\x)^2+s(\x)^2}\norm{{\bf f}(\x)}_2}{q(\x)}\right]\right)^2,
\end{align*}
}where $\norm{{\bf f}(\x)}_2$ denotes the euclidean norm.
The equality holds if and only if $\frac{\sqrt{m(\x)^2+s(\x)^2}\norm{{\bf f}(\x)}_2}{q(\x)}$ is constant.
Hence, the optimal proposal is
\begin{align}
	q_\text{opt}(\x)\propto \norm{{\bf f}(\x)}_2\sqrt{m(\x)^2+s(\x)^2}.
\end{align}

\subsection{Optimal proposal for $\widehat{{\bf I}}_{self}$}

Let us consider the case of the self-normalized estimator $\widehat{{\bf I}}_\text{self}$.
Recall that $\widehat{{\bf I}}_\text{self} = \frac{\widehat{{\bf E}}}{\widehat{Z}}$,
where $\widehat{{\bf E}}$ denotes the noisy estimator of ${\bf E} = \int_{\mathcal{X}} {\bf f}(\x)m(\x)d\x$, so that we are considering ratios of estimators.
Again, we aim to find the proposal that minimizes the variance of the vector-valued estimator $\widehat{{\bf I}}_\text{self}$.
When $N$ is large enough, the variance of $p$-th ratio is approximated  as \cite{libroStats},
\begin{align*}
\text{Var}[\widehat{I}_{\text{self},p}] = \text{Var}\left[\frac{\widehat{E}_p}{\widehat{Z}}\right] 
\approx \frac{1}{\bar{Z}^2}\text{Var}[\widehat{E}_p] - 2\frac{E_p}{Z}\text{Cov}[\widehat{E}_p,\widehat{Z}] + \frac{E_p^2}{\bar{Z}^4}\text{Var}[\widehat{Z}],
\end{align*}
where $E_p$ is the $p$-th component of ${\bf E}$, and
\begin{align*}
\text{Var}[\widehat{E}_p] &= \frac{1}{N}\mathbb{E}\left[\frac{f_p(\x)^2(m(\x)^2+s(\x)^2)}{q(\x)^2}\right] - \frac{1}{N}E_p^2, \\
\text{Var}[\widehat{Z}] &= \frac{1}{N}\mathbb{E}\left[\frac{m(\x)^2+s(\x)^2}{q(\x)^2}\right] - \frac{1}{N}\bar{Z}^2, \\
\text{Cov}[\widehat{E}_p,\widehat{Z}] &= \frac{1}{N}\mathbb{E}\left[\frac{f_p(\x)(m(\x)^2+s(\x)^2)}{q(\x)^2}\right] - \frac{1}{N}E_p\bar{Z}. 
\end{align*}
The first two results have been already obtained in the previous sections. 
The third result is given in Appendix \ref{mi_polla}.
The sum of the variances is thus
\begin{align*}
	\sum_{p=1}^{d_f}\text{Var}[\widehat{I}_{\text{self},p}] \approx \frac{1}{N\bar{Z}^2}\mathbb{E}\left[\frac{(m(\x)^2+s(\x)^2)\sum_{p=1}^{d_f}(f_p(\x) - I_p)^2}{q(\x)^2}\right].
\end{align*}
By Jensen's inequality, we can derive that the optimal proposal is 
\begin{align}
q_\text{opt}(\x) \propto \norm{{\bf f}(\x) - {\bf I}}_2\sqrt{m(\x)^2 + s(\x)^2}.	
\end{align}

\noindent
{\bf Relationship with active learning.} The optimal density $q_{opt}(\x)$ can be interpreted as an {\it acquisition density}, suggesting the regions of the space which require more number of acquisitions of the realizations $\widetilde{m}(\x)$. Namely, $q_{opt}(\x)$  plays a role similar to  an acquisition function in active learning. % remarking the regions which need more sampling.
 This is information is relevant, especially if the noisy evaluations are also costly to obtain.

%%%%%%%%%%%%%%%%%%%%%%%%%%%%%%
\subsection{Connection with other types of optimality}
%%%%%%%%%%%%%%%%%%%%%%%%%%%%%%

Here, we discuss another approach for optimality in noisy IS and connect it with our work. 
Other related works, in Monte Carlo and  noisy optimization literature,  focus on the trade-off between accuracy/noisiness and computational cost \cite{tran2013importance,doucet2015efficient,arnold2012noisy}.
In those settings, it is assumed that one can control the variance $s(\x)^2$ of the noisy realizations $\widetilde{m}(\x)$.
Clearly, taking samples with higher accuracy, i.e. small variance $s(\x)^2$, is beneficial since it decrease the magnitude of  the terms $\Exp\left[\frac{s(\x)^2}{m(\x)^2}\right]$ and $\Exp\left[\frac{f_p(\x)^2s(\x)^2}{m(\x)^2}\right]$, which are responsible for the efficiency loss in the estimators, due to the presence of noise. However, taking accurate estimates implies increased computational cost, hence one must reduce the number of samples $N$, which affect the overall Monte Carlo variance. This trade-off have been investigated in both MCMC and IS frameworks \cite{tran2013importance,doucet2015efficient}.

Let $R$ denote the number of auxiliary samples employed to reduce the variance of the noisy realizations. Namely, greater $R$ implies greater accuracy but also greater cost. Moreover, this number could depend on $\x$, i.e., $R(\x):\mathcal{X} \rightarrow \mathbb{N}^+\backslash\{0\}$. 
Then, the goal is to obtain the optimal function $R(\x)$ by balancing the decrease in variance with the extra computational cost (see, e.g., Sections 3.3, 3.4 and 5 of \cite{tran2013importance}). Namely, in this different approach, they try to reduce $s(\x)^2$ at certain $\x$ increasing the value of $R(\x)$, instead of using an optimal proposal pdf for the noisy scenario. On the contrary, in this work we have considered the use of optimal proposal pdfs and that $s(\x)^2$ is not tuned by the user, which means that $R$ is arbitrary and set constant for all $\x$.

\section{Numerical experiments}
\begin{figure}[h!]
	\centering
%\centerline{	
%	\begin{subfigure}[b]{0.5\textwidth}
%		\includegraphics[width=1\textwidth]{A_0_2.png}
%		\caption{$A=0.2$}
%	\end{subfigure}
%	\begin{subfigure}[b]{0.5\textwidth}
%		\includegraphics[width=1\textwidth]{A_0_5.png}
%		\caption{$A=0.5$}
%	\end{subfigure}
%	}
\centerline{	
%	\begin{subfigure}[b]{0.5\textwidth}
%		\includegraphics[width=1\textwidth]{A_1_2.png}
%		\caption{$A=1.2$}
%	\end{subfigure}
	\begin{subfigure}[b]{0.5\textwidth}
	\includegraphics[width=1\textwidth]{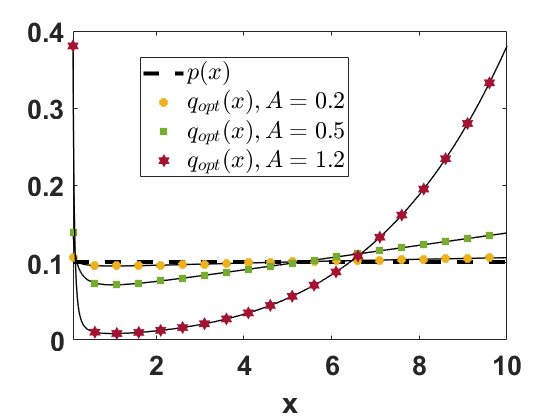}
	\caption{$q_\text{opt}(x)$ for different values of  $A$.}
\end{subfigure}
	\begin{subfigure}[b]{0.5\textwidth}
		\includegraphics[width=1\textwidth]{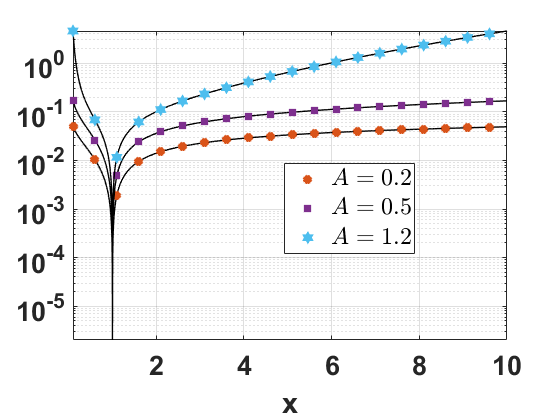}
		\caption{$s(x)$ for different values of  $A$ in log-scale.}
	\end{subfigure}
	}
	\caption{\label{no_me_toques_la_polla}	
		Uniform example.
		{\bf (a)} Optimal proposals $q_\text{opt}(x)$ for different values of $A$, and the $q_\text{sub-opt}(x)=p(x)$ in dashed line; {\bf (b)} The standard deviation $s(x)$ for different values of $A$.	
}
\end{figure}

\begin{figure}[h!]
	\centering
	\centerline{
		\begin{subfigure}[b]{0.5\textwidth}
			\includegraphics[width=1\textwidth]{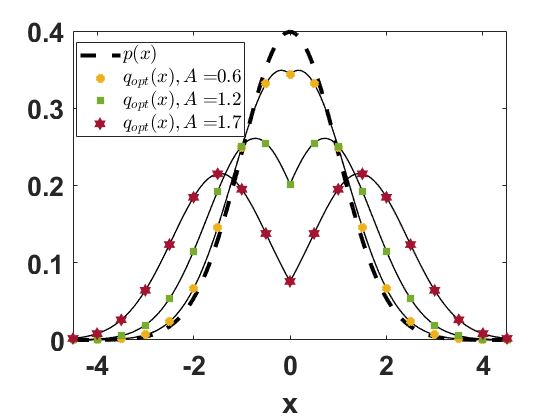}
			\caption{$q_\text{opt}(x)$ for different values of  $A$.}
		\end{subfigure}
		\begin{subfigure}[b]{0.5\textwidth}
			\includegraphics[width=1\textwidth]{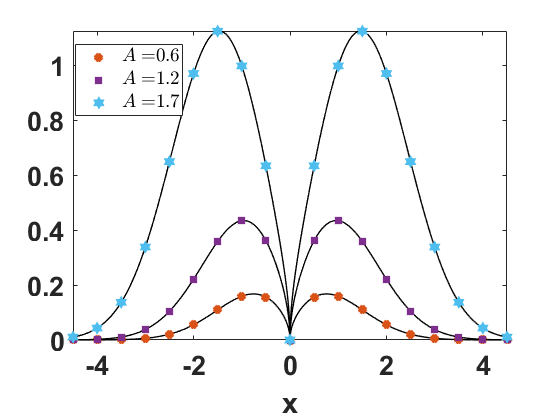}
			\caption{$s(x)$ for different values of  $A$.}
		\end{subfigure}
	}
	\caption{\label{fig_3}	Gaussian example.
		{\bf (a)} Optimal proposals $q_\text{opt}(x)$ for different values of $A$, and the $q_\text{sub-opt}(x)=p(x)$ in dashed line; {\bf (b)} The standard deviation $s(x)^2$ for different values of $A$.	
	}
\end{figure}

\begin{figure}[h!]
	\centering
	\centerline{
		\begin{subfigure}[b]{0.5\textwidth}
			\includegraphics[width=1\textwidth]{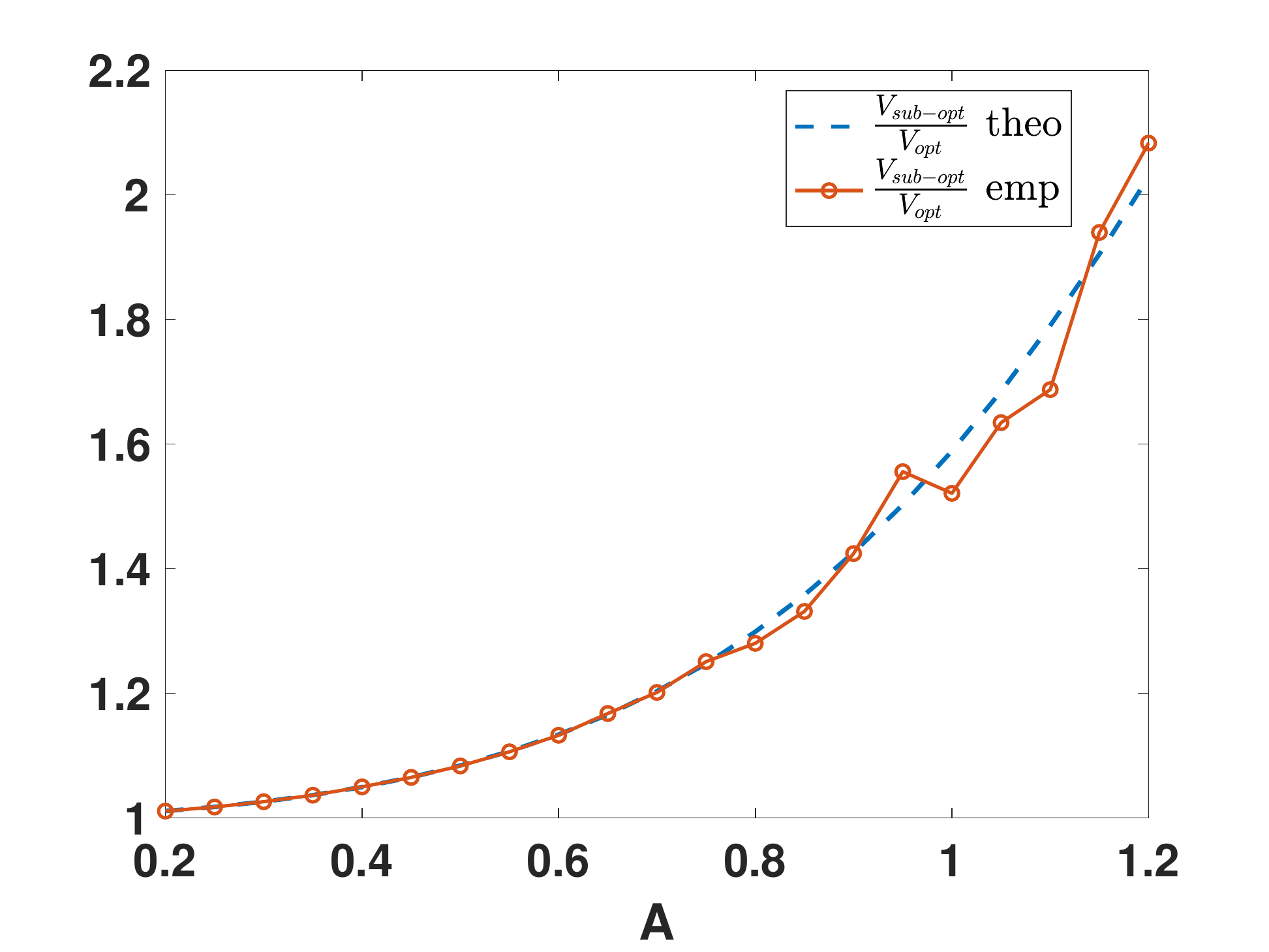}
			\caption{Uniform example.}
		\end{subfigure}
		\begin{subfigure}[b]{0.5\textwidth}
			\includegraphics[width=1\textwidth]{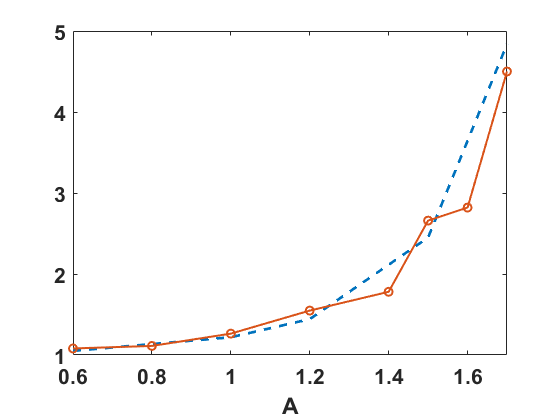}
			\caption{Gaussian example.}
		\end{subfigure}
	}
	\caption{\label{fig_2}	
		Theoretical and empirical ratio of variances $\frac{V_\text{sub-opt}}{V_\text{opt}}$ in the estimation of $\bar{Z}=1$ for both experiments. The $x$-axis denotes the noise level (larger $A$ means greater noise).	
	}
\end{figure}

%%%%%%%%%%%%%%%%%%%%%%%%%%%%%%%%%%%%
In this section, we consider two illustrative numerical example where we clearly show the performance of the optimal proposal pdf in the noisy IS setting (showing the variance gains in estimation, with respect to the use the optimal proposal density from the non-noisy setting). For simplicity,  we consider one-dimensional scenarios, and test the optimal proposal pdf with different densities $p(x)$ (uniform and Gaussian), and different types of variance behavior, $\sigma(x)$.
\newline
\newline
{\bf First experiment.} Let $p(x) = \frac{1}{b-a}$ for $x \in [a,b]$, i.e., a uniform density in $[a,b]$ with $a=0.1$ and $b=10$. We set $\widetilde{m}(\x) = p(\x)e^\epsilon$ with $\epsilon \sim \mathcal{N}\left(-{\sigma^2}/{2},\sigma^2\right)$ so that $\Exp[e^\epsilon]=1$, and we have $m(x)=\mathbb{E}[\widetilde{m}(\x) ]=p(x)$.
\newline
We consider the estimation of $\bar{Z}=1$ using the optimal proposal pdf $q_\text{opt}(x) $ in Eq. \eqref{aquiVaOpt}, and the optimal proposal pdf in the non-noisy setting, i.e., $q_\text{sub-opt}(x)=p(x)$.
More specifically,  we consider
$$
 \sigma(x)= A|\log(x)|, \quad A>0.
$$
Hence,
$$
s(x)^2 = \frac{e^{\sigma(x)^2}-1}{(b-a)^2},\quad  \text{and} \quad q_\text{opt}(x) \propto \frac{1}{b-a}e^{\sigma(x)^2}.
%\exp\left(\sigma(x)^2/2\right).
$$
Clearly, by changing $A$, we change the form of both $s(x)^2$ and $q_\text{opt}(x)$. 
For instance, increasing $A$ also increases the magnitude of $s(x)^2$ and hence the mismatch between $q_\text{sub-opt}(x)=p(x)$ and $q_\text{opt}(x)$, as depicted in Figure \ref{no_me_toques_la_polla}.
Indeed, for $A=0.2$, $q_\text{opt}(x)$ is almost identical to $p(x)$ since the magnitude of $s(x)$ is small w.r.t. the values of $p(x)$. As $A$ increases, $q_\text{opt}(x)$ deviates from $p(x)$, being in the middle between $p(x)$ and $s(x)$, and eventually would converge to $s(x)$ for $A\gg 1$. 
It is also interesting to note that $q_\text{opt}(x)$ with $A=1.2$ has very little probability mass around $x=1$, where the noise is zero, since it needs to concentrate probability mass in the extremes of the interval, where the noise power is huge.
\newline
Let also denote as $V_\text{sub-opt}$ the variance obtained using $q_\text{sub-opt}(x)=p(x)$ given in  Eq. \eqref{eq_Vopt}, and $V_\text{opt}=V_{\text{min}}$   the variance obtained using $q_\text{opt}(x)$ given in Eq.  \eqref{eq_Vsub}.
In Figure \ref{fig_2}(a), we show the ratio of variances $\frac{V_\text{sub-opt}}{V_\text{opt}}$ both theoretically and empirically, as a function of $A$, where $V_\text{sub-opt}$ and $V_\text{opt}$ are the variances of $\widehat{Z}$ when using $p(x)$ and $q_\text{opt}(x)$ as proposals, respectively. We can observe the clear advantage of using the optimal proposal density $q_\text{opt}(x) $ in Eq. \eqref{aquiVaOpt}.
\newline
\newline
{\bf Second experiment.} 
Let us consider now a Gaussian pdf, $p(x) = \mathcal{N}(x|0,1)$, and the same error model as in the previous example but considering 
$$
\sigma(x)= A|x|^\frac{1}{2}, \quad A>0.
$$
Figure \ref{fig_3} depicts the $q_\text{opt}(x)$ and $s(x)$, as a function of $x$, for several values of $A$. Note that, in this example, increasing $A$ makes $q_\text{opt}(x)$ become bimodal. 
As in the previous example, as $A$ increases, the optimal proposal $q_\text{opt}(x)$ will converge to $s(x)$.
The theoretical and empirical curves of the ratio of variances, $\frac{V_\text{sub-opt}}{V_\text{opt}}$, in estimating $Z$ are shown in Figure \ref{fig_2}(b).

%\begin{figure}[h!]
%	\centering
%	\includegraphics[width=0.5\textwidth]{ratio_teo_y_emp.pdf}
%	\caption{\label{fig_2}	
%		Ratio of variances $\frac{V_\text{sub-opt}}{V_\text{opt}}$.	
%	}
%\end{figure}

%%%%%%%%%%%%%%%%%%%%%%%%%
%%%%%%%%%%%%%%%%%%%%%%%%%
\section{Conclusions}
\label{sec:conclusions}
%%%%%%%%%%%%%%%%%%%%%%%%%
%%%%%%%%%%%%%%%%%%%%%%%%%

 Working with noisy evaluations of the target density is usual in Monte Carlo, especially in the last years. In this work, we have analyzed the use of optimal proposal densities in a noisy IS framework. Previous works have focused on the trade-off between accuracy in the evaluation and computational cost in order to form optimal estimators. In this work, we have considered a general setting and derived the optimal proposals for the noisy IS estimators. These optimal proposals incorporate the variance function of the noisy evaluation in order to propose samples in regions that are more affected by noise. In this sense, we can informally state that the optimal proposal densities play the role of an acquisition function that also take into account the noise power.

\bibliographystyle{IEEEbib}
\bibliography{bibliografia}

\begin{appendices}
	\section{Covariance between $\widehat{E}_p$ and $\widehat{Z}$}\label{mi_polla}
	We show that
	$$
	\text{Cov}[\widehat{E}_p,\widehat{Z}] = \frac{1}{N}\mathbb{E}\left[\frac{f_p(\x)(m(\x)^2+s(\x)^2)}{q(\x)^2}\right] - \frac{1}{N}E_p\bar{Z}.
	$$
	First, recall that $\text{Cov}[\widehat{E}_p,\widehat{Z}] = \Exp[\widehat{E}_p\widehat{Z}] - E_p\bar{Z}$.
	By the law of iterated expectations,
	\begin{align*}
		\Exp[\widehat{E}_p\widehat{Z}] = \Exp[\Exp[\widehat{E}_p\widehat{Z}|\x_{1:N}]].
	\end{align*}
	The inner expectation is 
	{\footnotesize\begin{align*}
		\Exp[\widehat{E}_p\widehat{Z}|\x_{1:N}] 
		&=  \Exp\left[\frac{1}{N^2}\sum_{i=1}^{N}w_i^2f_p(\x_i) + \frac{2}{N^2}\sum_{i=1}^N\sum_{j>i}^{N}w_iw_jf_p(\x_i) \bigg| \x_{1:N}\right] \\
		&= \frac{1}{N^2}\sum_{i=1}^{N}\frac{f_p(\x_i)(s(\x_i)^2+m(\x_i)^2)}{q(\x_i)^2} + \frac{2}{N^2}\sum_{i=1}^N\sum_{j>i}^{N}\frac{m(\x_i)f(\x_i)}{q(\x_i)}\frac{m(\x_j)}{q(\x_j)}.
	\end{align*}}
	Hence, we obtain
	{\footnotesize\begin{align*}
		\Exp\left[\Exp[\widehat{E}_p\widehat{Z}|\x_{1:N}]\right] &= \frac{1}{N}\Exp\left[\frac{f(\x)(s(\x)^2+m(\x)^2)}{q(\x)^2}\right] \\
		&+ \frac{2}{N^2}\sum_{i=1}^N\sum_{j>i}^{N}\Exp\left[\frac{m(\x_i)f(\x_i)}{q(\x_i)}\right]\Exp\left[\frac{m(\x_j)}{q(\x_j)}\right] \\
		&=\frac{1}{N}\Exp\left[\frac{f(\x)(s(\x)^2+m(\x)^2)}{q(\x)^2}\right] +
		\frac{2}{N^2}\sum_{i=1}^N\sum_{j>i}^{N}E_p\bar{Z} \\
		&=\frac{1}{N}\Exp\left[\frac{f(\x)(s(\x)^2+m(\x)^2)}{q(\x)^2}\right] +
		E_p\bar{Z}\left(1-\frac{1}{N}\right).
	\end{align*}}
	Combining the results, we obtain the desired expression.

\end{appendices}

\end{document}